\documentclass{article}

\usepackage{amssymb}
\usepackage{amsmath}
\usepackage{stmaryrd}
\usepackage[margin=1in]{geometry}
\usepackage{graphics}
\usepackage{tikz}
\usepackage{epsfig}
\usepackage{multirow}
\usepackage{graphicx}
\usepackage[center]{caption}
\usepackage{subcaption}
\usepackage{cancel}
\usepackage[hidelinks]{hyperref}
\usepackage{graphicx} 
\title{Fitting magnetization data using continued fraction of straight lines}
\author{Vijay Prakash S.,\\
	\small Independent Researcher, Alappuzha, Kerala, India.\\
	\small \tt{prakash.vijay.s@gmail.com}}
\date{}
\begin{document}
	\maketitle
	\begin{abstract}
		\noindent
Magnetization of a ferromagnetic substance in response to an externally applied magnetic field increases with the strength of the field. This is because at the microscopic level, magnetic moments in certain regions or domains of the substance increasingly align with the applied field, while the amount of misaligned domains decreases. The alignment of such magnetic domains with an applied magnetic field forms the physical basis for the nonlinearity of magnetization. In this paper, the nonlinear function is approximated as a combination of continued fraction of straight lines. 
The resulting fit is used to interpret the nonlinear behavior in both growing and shrinking magnetic domains. The continued fraction of straight lines used here is an algebraic expression which can be used to estimate parameters using nonlinear regression.
	\end{abstract}
	\section{Introduction}
	\label{sec_intro}
From single-molecule magnets \cite{smm} to large-scale synchrotron magnetic units \cite{synchrotron}, magnetization becomes a major factor for various scales of applications. Of particular interest, is the magnetization of a ferromagnetic substance which has also been one of the oldest topics of scientific research \cite{oldest}-\cite{oldest3}. 
Magnetization of this kind is usually analyzed with the following plots:
\begin{enumerate}
\item basic (i.e. initial or first) magnetization curves \cite{basic0} - \cite{basic2},
\item alternating magnetization curves with hysteresis loop \cite{mdpi_23}, and
\item demagnetization curves \cite{demag0} - \cite{demag1}.
\end{enumerate}
Many models have been proposed to trace the above set of curves. Most of them are based on differential equations \cite{mdpi_23}. Some modeling approaches involve numerical methods such as finite element methods \cite{demag01,fem0,fem,alnico} and data-driven updating \cite{datad} including neural networks with sequence-to-sequence architecture \cite{nn1}. 

Other than the above methods, analytical models have also been proposed to trace the above curves. These include polynomials, functions of exponential, trigonometric or sigmoidal nature, and monotocity preserving splines \cite{mdpi_23,analy1,analy2,analy3}. These methods assume that the above set of curves as nonlinear functions of the $x-$ axis, i.e., magnetization as a nonlinear function of the applied magnetic field.

Instead of assuming a fixed form of nonlinearity, in this work, we introduce nonlinearity through $y-$ axis parametrically that takes the form of continued fraction of straight lines \cite{1st_paper}. The magnetization that increases under applied magnetic field and reaches saturation, is expressed as a continued fraction. The parametric form generalizes the model making it useful for fitting on data.

\section{Magnetization and dissipation}
Magnetization with nonlinear dissipation can be expressed using continued fraction of $y-y_c=m(x-x_c)$ (with slope $m$) about a point $(x_c,y_c)$ given by \cite{1st_paper,a_m_mod}
\begin{eqnarray}
\nonumber y-y_c &=& \frac{m(x-x_c)}{1+a(y-y_c)^2}\\
\Rightarrow y-y_c &=& \frac{m(x-x_c)}{1+a\frac{m^2(x-x_c)^2}{\left( 1+a\left( \frac{m(x-x_c)}{1+\cdots}\right)^2\right)^2}},
\label{eq_contfrac}
\end{eqnarray}
where $a>0$ is a parameter that provides a positive nonlinear damping to magnetization.
The above equation is the algebraic expression: $a(y-y_c)^3+(y-y_c)=m(x-x_c)$ which has the solution $y=S_1+S_2+y_c$, where
\begin{eqnarray}
\nonumber S_1&=& \frac{-1}{3}\left[-\left(\frac{27m(x-x_c)}{2a}\right)+\sqrt{\left(\frac{27m(x-x_c)}{2a}\right)^2+\frac{27}{a^3}}\right]^{{1}/{3}}\ \text{and}\\
S_2&=& \frac{1}{a}\left[-\left(\frac{27m(x-x_c)}{2a}\right)+\sqrt{\left(\frac{27m(x-x_c)}{2a}\right)^2+\frac{27}{a^3}}\right]^{{-1}/{3}}.
\label{eq_s1s2}
\end{eqnarray} 
Due to continued fraction, the straight line becomes an S-curve with inflection at the point $(x_c,y_c)$. $m$ is now the maximum slope value of the S-curve representing maximum permeability of a ferromagnetic material. 
\subsection{A classical example}
\begin{figure}[ht!]
	\centering
	\begin{subfigure}[b]{0.45\textwidth}
		\includegraphics[width=\textwidth]{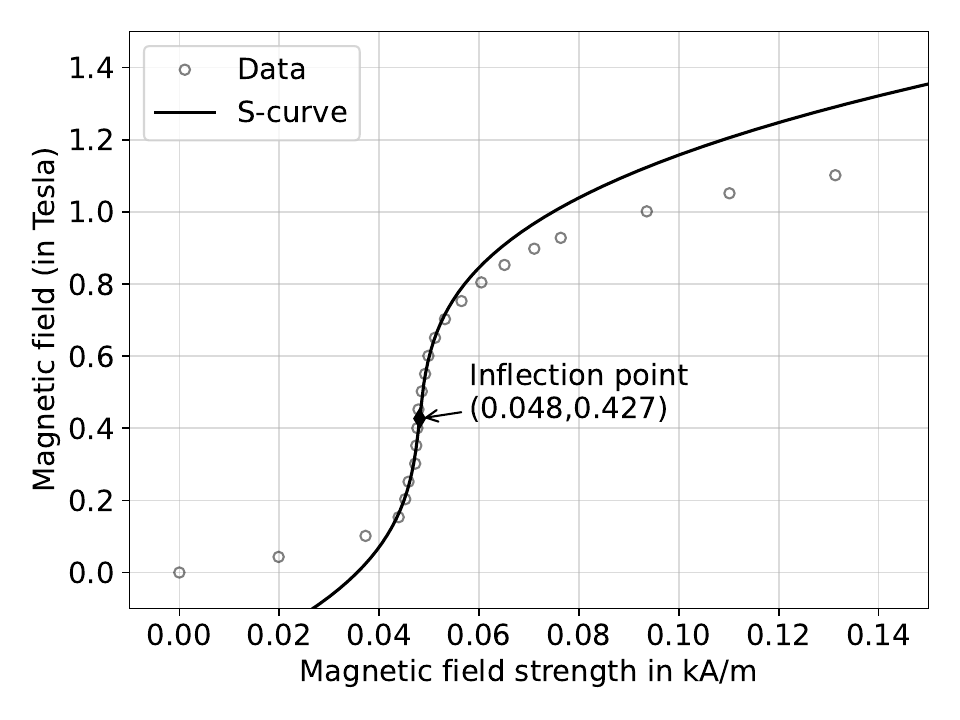}
		\caption{$m=0.131 H/m$ and $a=15.52 T^{-2}$. Fitting $y$ from fourth to fifteenth data point with the point of inflection as the mean.}
		\label{fig1a_am}
	\end{subfigure}
	\begin{subfigure}[b]{0.45\textwidth}
		\includegraphics[width=\textwidth]{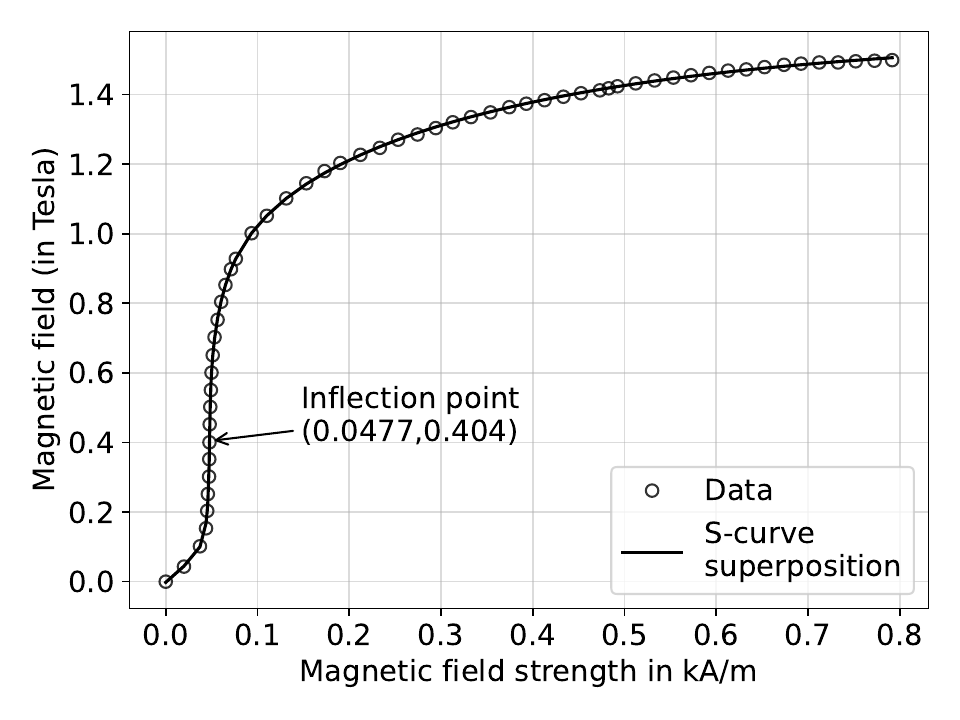}
		\caption{Maximum permeability $= 0.163 \text{H/m}$ and $a=350.03 \text{T}^{-2}$. Fitting $y_\text{net}$ over entire data by superposing 7 S-curves ($n=7$).}
		\label{fig1b_amsum}
	\end{subfigure}
	\caption{Fitting two-parameter S-curve and its superposition on magnetization data iron of commercial purity grade \cite{irodov}. The data points are obtained using WebPlotDigitilizer \cite{webplot}. S-curve superposition provides accurate estimation of inflection point and permeability at that point. Parameter $a$ varies according to the fitting conditions and reflects reduction in permeability as the curve deviates from the point of inflection \cite{twosupS}.}.
	\label{fig1_basic0}
\end{figure}

We now consider a classical basic magnetization curve (Fig. 3.76 of \cite{irodov}) of iron and fit $y(x,a,m,x_c,y_c)$ following the method in \cite{a_m_mod} which involves fitting over a selected set of data points. The result, shown in Fig. (\ref{fig1a_am}), fits partially around the inflection point. This indicates the relationship of magnetization with the applied field is more nonlinear and we tackle this with a combination of multiple S-curves \cite{twosupS}. We now fit the following superposition
\begin{eqnarray}
\nonumber y_{\text{net}} &=& \qquad\sum_{i=0}^{n-1} p_iy(a,m_i,x-x_{ci},y_{ci})\\
\Rightarrow y_{\text{net}} &=& \qquad\sum_{i=0}^{n-1} p_i\left(S_1(a,m_i,x-x_{ci})+S_2(a,m_i,x-x_{ci})+y_{ci}\right),
\label{eq_ynet}
\end{eqnarray}
where $n$ is the number of S-curves, $p_i$'s are weights, $m_i$ is the $i^{\text{th}}$ slope with $x_{ci}$ and $y_{ci}$ as the coordinates of inflection points. Fig. (\ref{fig1b_amsum}) shows the fit of S-curve superposition model.
\subsection{Permeability and dissipativity}
\begin{figure}[ht!]
	\centering
	\begin{subfigure}[b]{0.45\textwidth}
		\includegraphics[width=\textwidth]{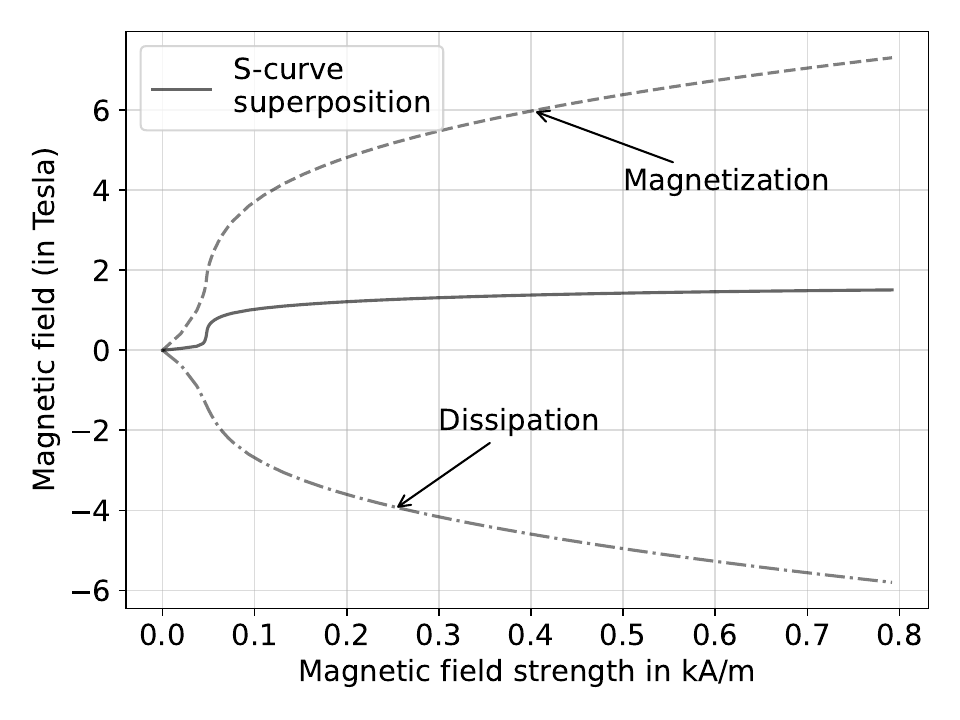}
		\caption{Magnetization and dissipation curves influence each other through the parameter $a$.}
		\label{fig2a}
	\end{subfigure}
	\begin{subfigure}[b]{0.45\textwidth}
		\includegraphics[width=\textwidth]{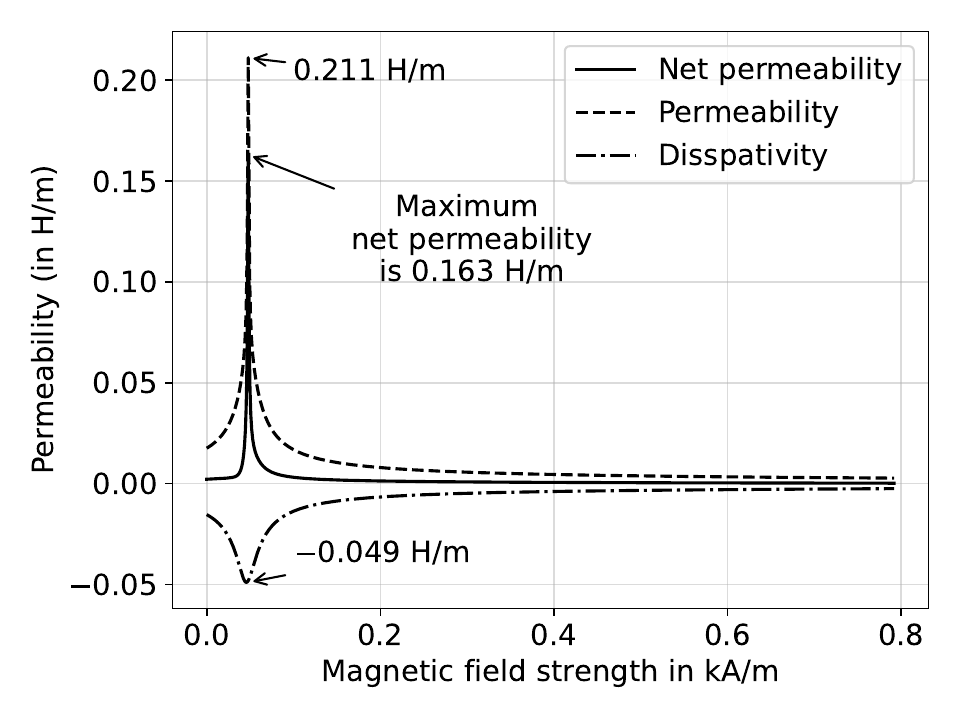}
		\caption{Both permeability and dissipativity reach highest value at the point of inflection.}
		\label{fig2b}
	\end{subfigure}
	\caption{Approximations due to positive and negative slope values $p_im_i$ from Eqn. (\ref{eqn_sup}) are shown separately. Permeability and dissipativity represent positive and negative values of $p_im_i$. They influence each other through $a$. Continued fraction of permeability is due to dissipativity and vice-versa. }.
	\label{fig2}
\end{figure}
In the previous section, from Fig. (\ref{fig1_basic0}) it can be seen that $y_\text{net}$ fits the magnetization data better than $y$. However, $y_\text{net}$ has significantly more number of parameters. To understand the superposition model, Eqn. (\ref{eq_ynet}) is rewritten as
\begin{eqnarray}
\nonumber y_{\text{net}} &=& \qquad\sum_{i=0}^{n-1} p_iy_i(x,a,m_i,x_{ci},y_{ci})\\
\Rightarrow y_{\text{net}} &=& \qquad\sum_{i=0}^{n-1} p_i\left(\frac{m_i(x-x_{ci})}{1+a(y-y_{ci})^2}+y_{ci}\right).
\label{eqn_sup}
\end{eqnarray}
The slopes of individual S-curves are given by $p_im_i$. Slopes $p_im_i\ge0$ represents permeability while those with $p_im_i<0$ are slopes of dissipative S-curves. Hence, we use the term `dissipativity' to refer to negative slopes in the superposition. Figs. (\ref{fig2a}) and (\ref{fig2b}) show the magnetizing and dissipating components, and their variations with the applied field strength, respectively.
\subsection{Magnetization subprocesses}
Combinations of the two components of the superposition model are useful in identifying the two subprocesses of magnetization. Using continued fraction of straight lines, an S-curve can be expressed as a sum of two distinct S-curves \cite{shruti_Ssum}. The S-curve governed by $y_\text{net}$ can be written as
\begin{equation}
y_{\text{net}}=S_I +S_{II}+\underbrace{\sum_{i=0}^{n-1} p_i y_{ci}}_{\text{offset}},
\label{eq_am_s}
\end{equation}
where $S_I = \sum_{i=0}^{n-1} p_iS_1$ and $S_{II} = \sum_{i=0}^{n-1} p_iS_2$ using Eqn. (\ref{eq_s1s2}).
$S_I$ and $S_{II}$ represent two subprocesses of the magnetization process $y_{\text{net}}$.
\begin{figure}[ht!]
	\centering
	\begin{subfigure}[b]{0.45\textwidth}
		\includegraphics[width=\textwidth]{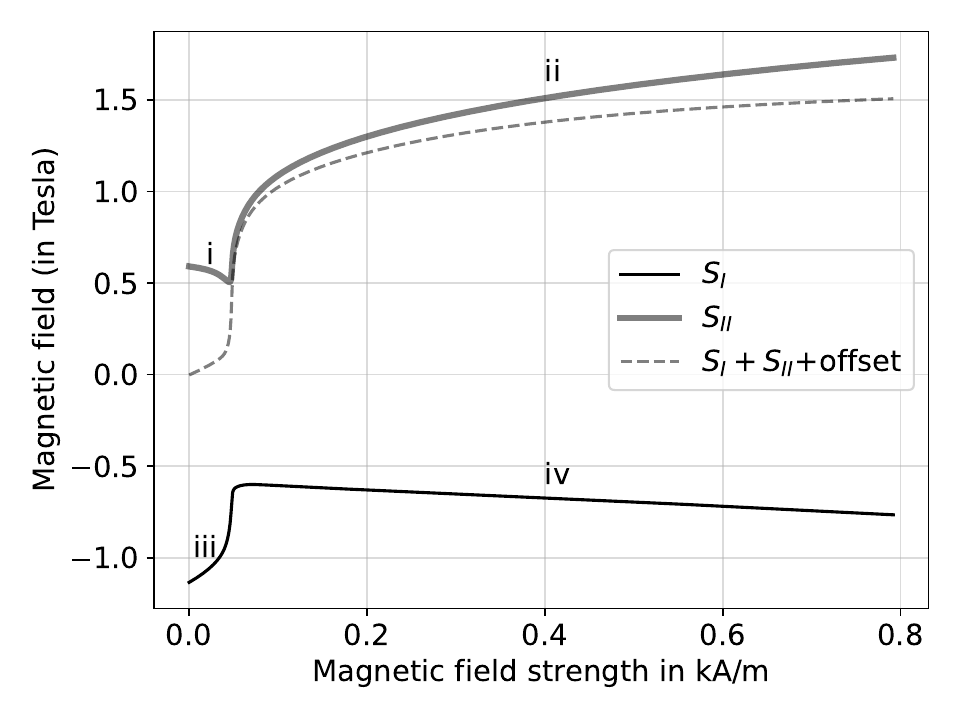}
		\caption{The subprocesses $S_I$ and $S_{II}$ in a magnetization curve are shown.}
		\label{fig3a}
	\end{subfigure}
	\begin{subfigure}[b]{0.45\textwidth}
		\includegraphics[width=\textwidth]{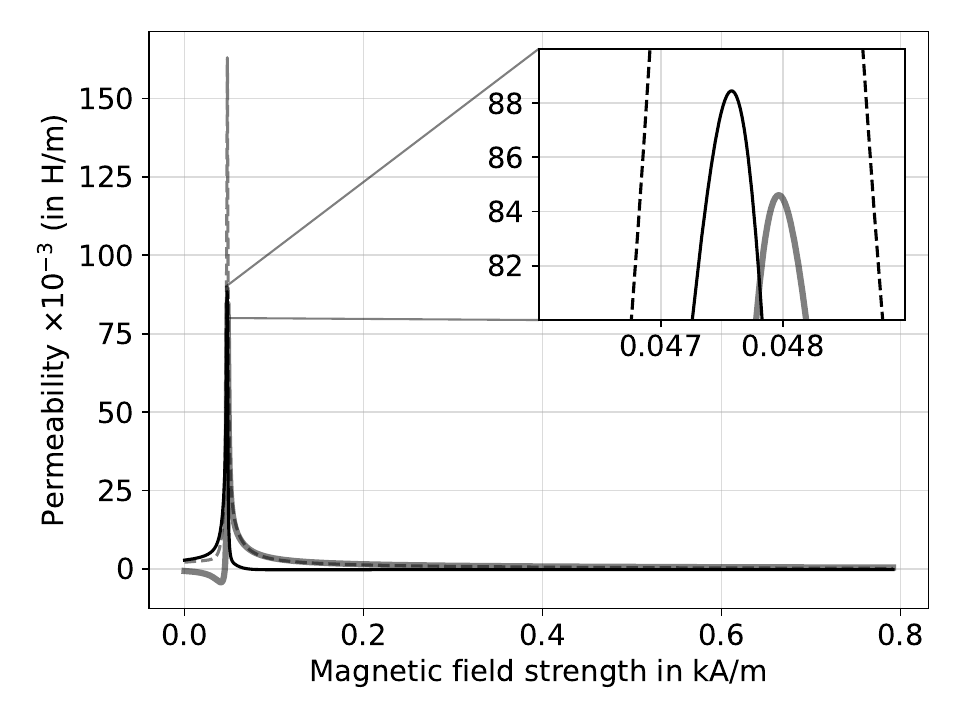}
		\caption{Derivatives of $S_I$ and $S_{II}$ along with an inset figure showing successive peaks.}
		\label{fig3b}
	\end{subfigure}
	\caption{With the applied magnetic field, (a) $S_I$ switches from magnetization dominant behavior to dominantly dissipative behavior, on the other hand, $S_{II}$ becomes magnetization dominant in the later part; (b) $S_I$ reaches maximum permeability followed by $S_{II}$.}
	\label{fig3}
\end{figure}
The two subprocesses in Fig. (\ref{fig3a}) represent the magnetization in different domains of the ferromagnetic substance. The curve $S_I$ represents domains with magnetic moments that are already aligned with the externally applied magnetic field. The region (iii) of $S_I$ in Fig. (\ref{fig3a}) reflects the increase in magnetization of these domains with the applied magnetic strength. After the aligned domains reach peak magnetization, dissipation becomes dominant and increases in region (iv).

$S_{II}$ in Fig. (\ref{fig3a}) represents domains of misaligned magnetic moments with the external magnetic field. (i) of  $S_{II}$ shows decrease in magnetization due to domain wall motion \cite{domain0,domain}. However, magnetization increases in (ii) in Fig. (\ref{fig3a}) after reaching a minimum as domains get increasingly aligned with the external field and reaches saturation. 

The successive peaks in Fig. (\ref{fig3b}) shows that the domains of aligned magnetic moments reach peak permeability followed by initially misaligned domains. The inflection point or the point of maximum permeability of the basic magnetization curve lies in between the successive peaks.
\section{Hysteresis loop}
\begin{figure}[ht!]
	\centering
 	\begin{subfigure}[b]{0.45\textwidth}
		\includegraphics[width=\textwidth]{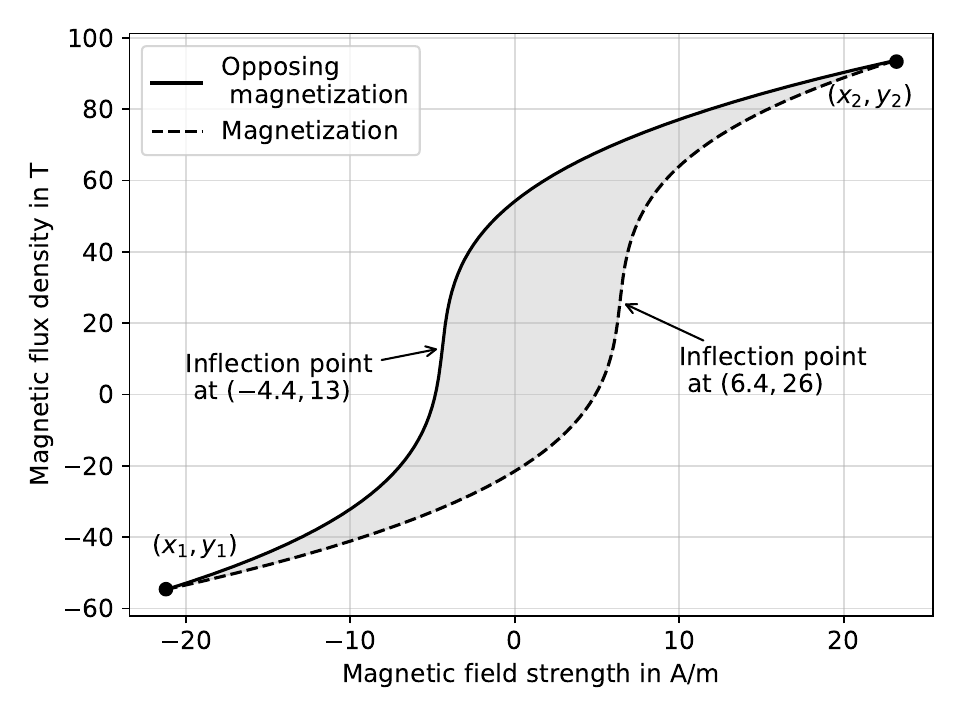}
		\caption{Representative plot of hysteresis loop with $a=0.002,\ m=41$.}
		\label{fig4a}
	\end{subfigure}
	\begin{subfigure}[b]{0.45\textwidth}
		\includegraphics[width=\textwidth]{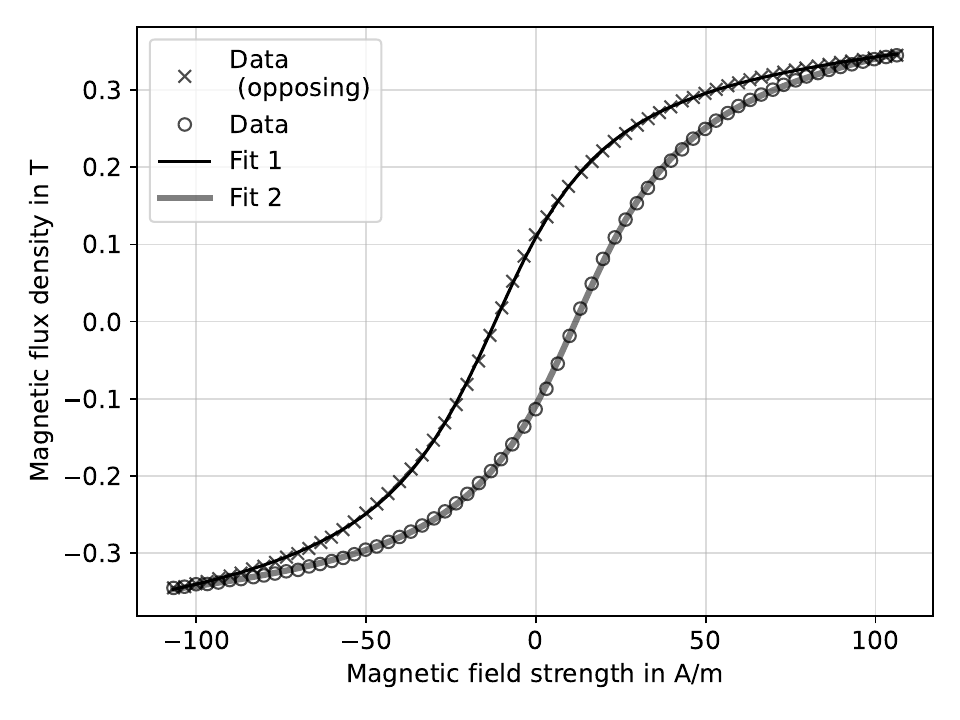}
		\caption{Hysteresis data exhibited by Mn-Zn ferrite \cite{maganalyst,anhys5} fitted with 2 S-curves ($n=2$).}
		\label{fig4b}
	\end{subfigure}
	\caption{Hysteresis loop shown in (a) with only the magnetization component.(b) data approximation using  $y_{\text{net}}$ with $n=2$ including both the magnetization and dissipation components. The upper and lower curve in (b) intersect at $(-106.093,-0.346)$ and $(105.503, 0.346)$. The area of the loop in (b) is 14.74 J.}
	\label{fig4}
\end{figure}
\begin{figure}[ht!]
	\centering
	\begin{subfigure}[b]{0.45\textwidth}
		\includegraphics[width=\textwidth]{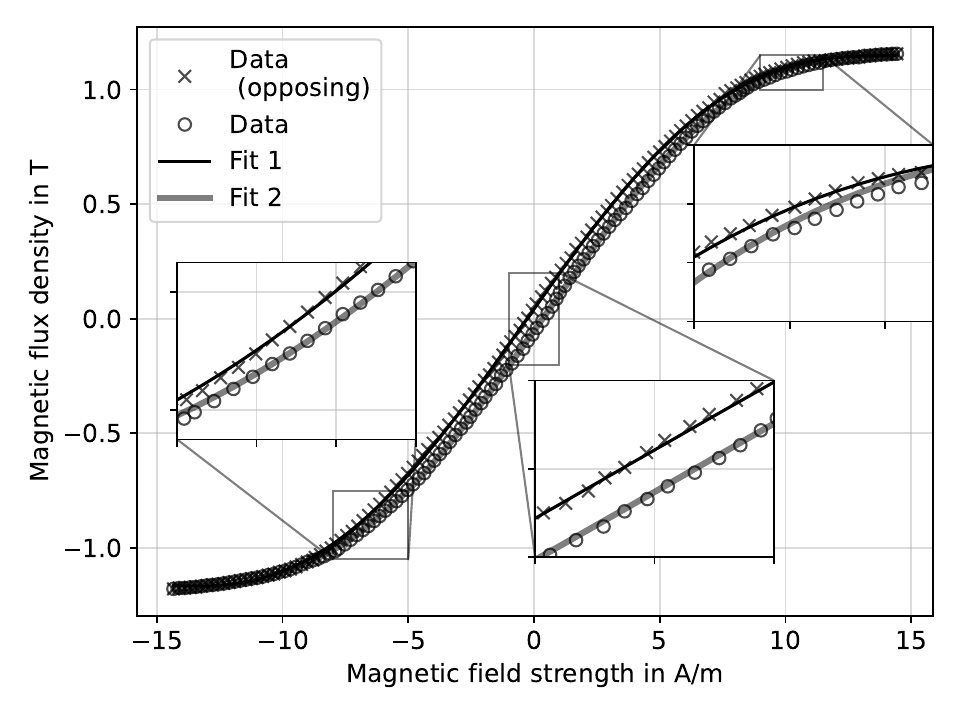}
		\caption{Hysteresis behavior of Fe$_{73.5}$Si$_{13.5}$B$_9$Nb$_3$Cu$_1$ nanocrystalline alloy \cite{anhys6} fitted with 7 S-curves $(n=7)$.}
		\label{fig5a}
	\end{subfigure}
	\begin{subfigure}[b]{0.45\textwidth}
		\includegraphics[width=\textwidth]{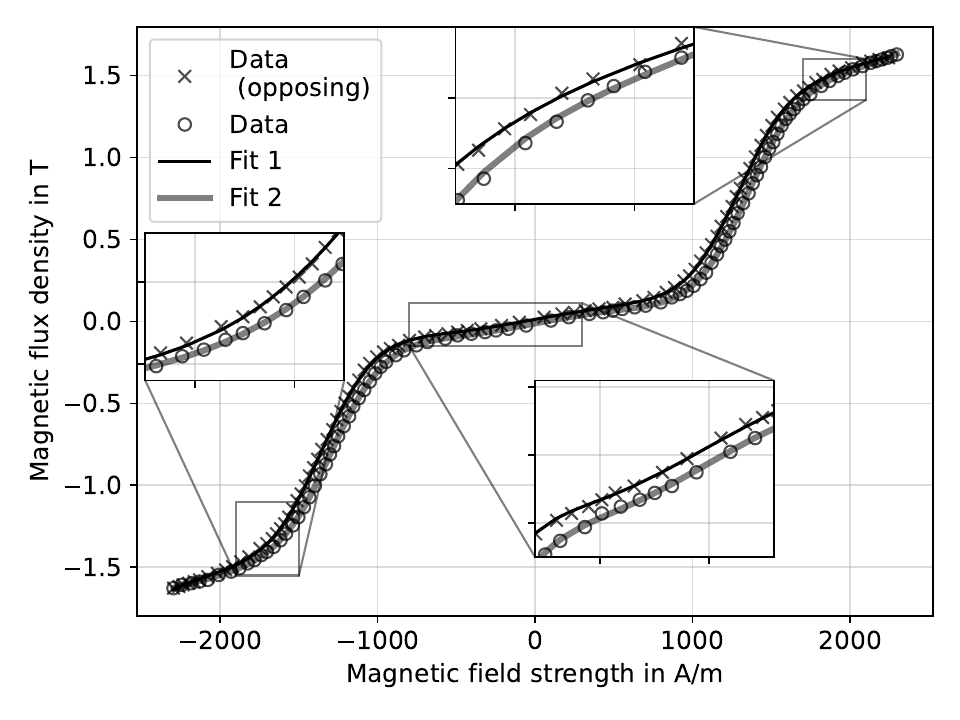}
		\caption{Hysteresis behavior of grain-oriented silicon steel \cite{anhys6} fitted with 7 S-curves $(n=7)$.}
		\label{fig5b}
	\end{subfigure}
	\caption{Hysteresis behavior of soft magnets \cite{maganalyst}. (a) The estimated end points of hysteresis loop are ($-13.126$ A/m,$-1.167$ T) and (16.456 A/m, 1.166 T) that enclose an area of 1.1945 J.(b) The estimated end points of hysteresis loop are ($-2247.3$ A/m,$-1.614$ T) and (2248.9 A/m, 1.614 T) that enclose an area of 243.52 J.}
	\label{fig5}
\end{figure}
\begin{figure}[ht!]
	\centering
	\begin{subfigure}[b]{0.45\textwidth}
		\includegraphics[width=\textwidth]{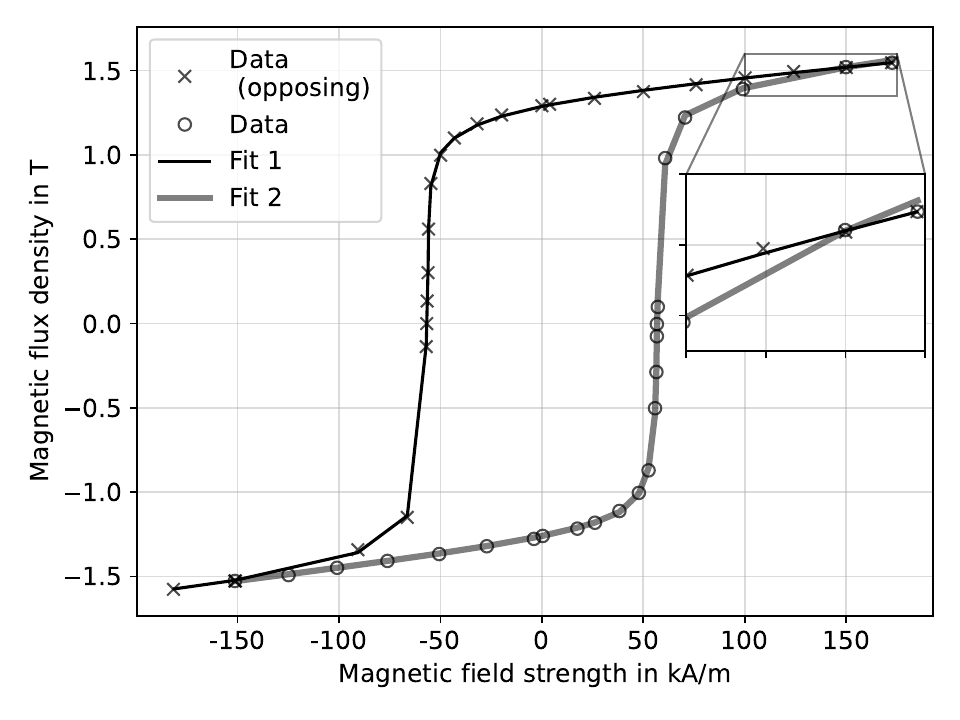}
		\caption{Hysteresis behavior of Alnico 5 \cite{alnico00} fitted with 5 S-curves $(n=5)$.}
		\label{fig6a}
	\end{subfigure}
	\begin{subfigure}[b]{0.45\textwidth}
		\includegraphics[width=\textwidth]{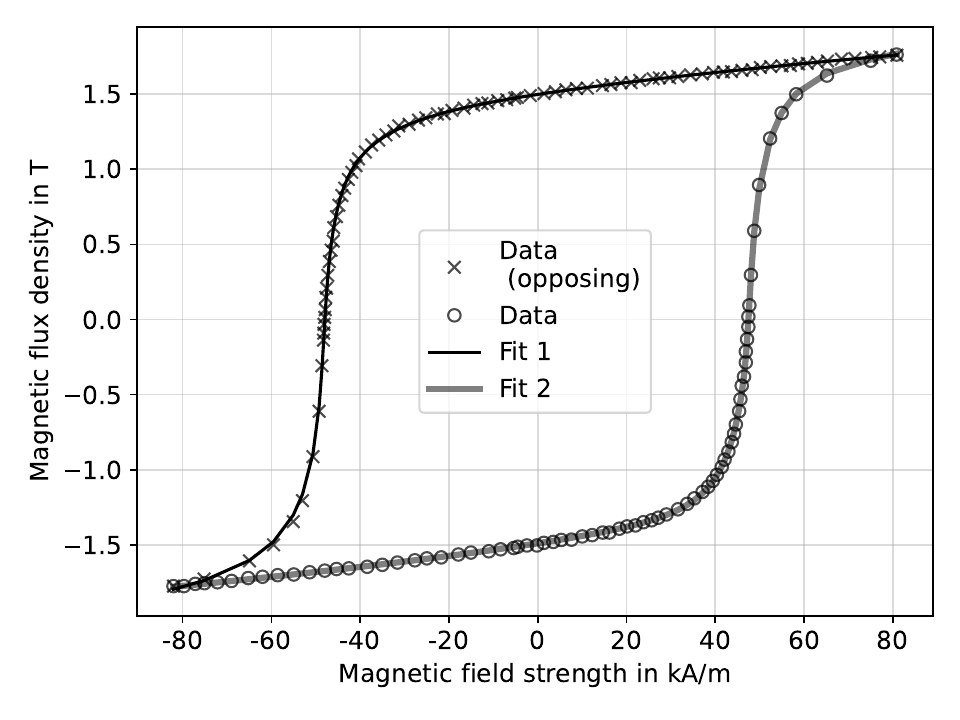}
		\caption{Hysteresis behavior of Alnico \cite{alnico0} fitted with 3 S-curves $(n=3)$.}
		\label{fig6b}
	\end{subfigure}
	\caption{Hysteresis behavior of Alnico \cite{hyster_data}. (a) The estimated end points of hysteresis loop are ($-175.15$ kA/m,$-1.564$ T) and (150.18 kA/m, 1.52 T) that enclose an area of 299.93 kJ.(b) The estimated end points of hysteresis loop are ($-78.76$ kA/m,$-1.77$ T) and (78.53 kA/m, 1.755 T) that enclose an area of 287.46 kJ.}
	\label{fig6}
\end{figure}
\begin{figure}[ht!]
	\centering
	\begin{subfigure}[b]{0.45\textwidth}
		\includegraphics[width=\textwidth]{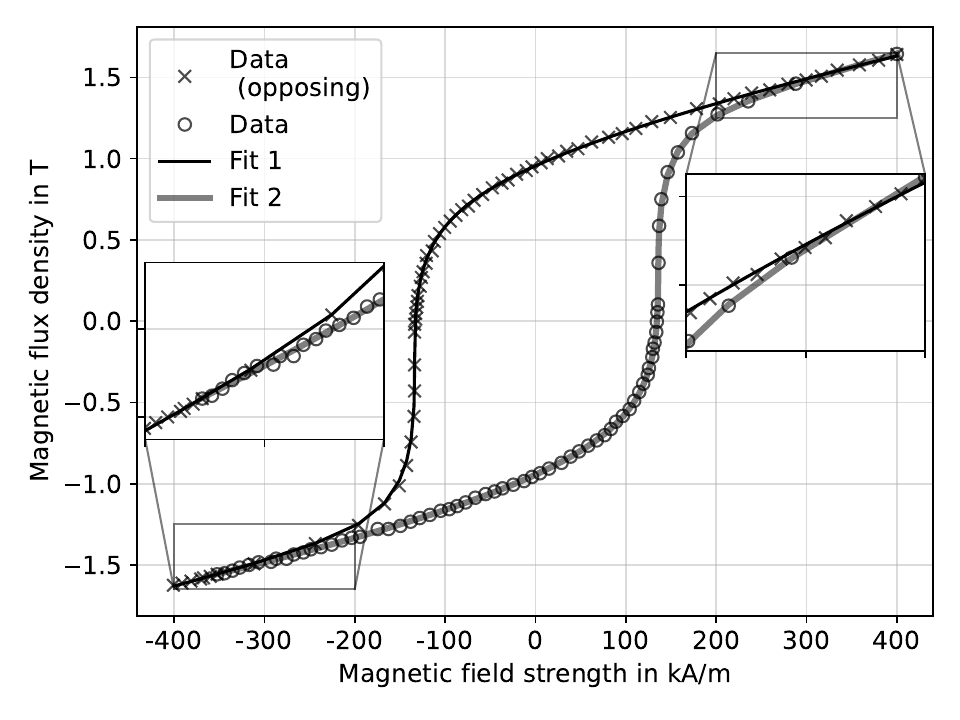}
		\caption{Hysteresis behavior of Alnico 8 LNGT40 \cite{alnico} fitted with 4 S-curves.}
		\label{fig7a}
	\end{subfigure}
	\begin{subfigure}[b]{0.45\textwidth}
		\includegraphics[width=\textwidth]{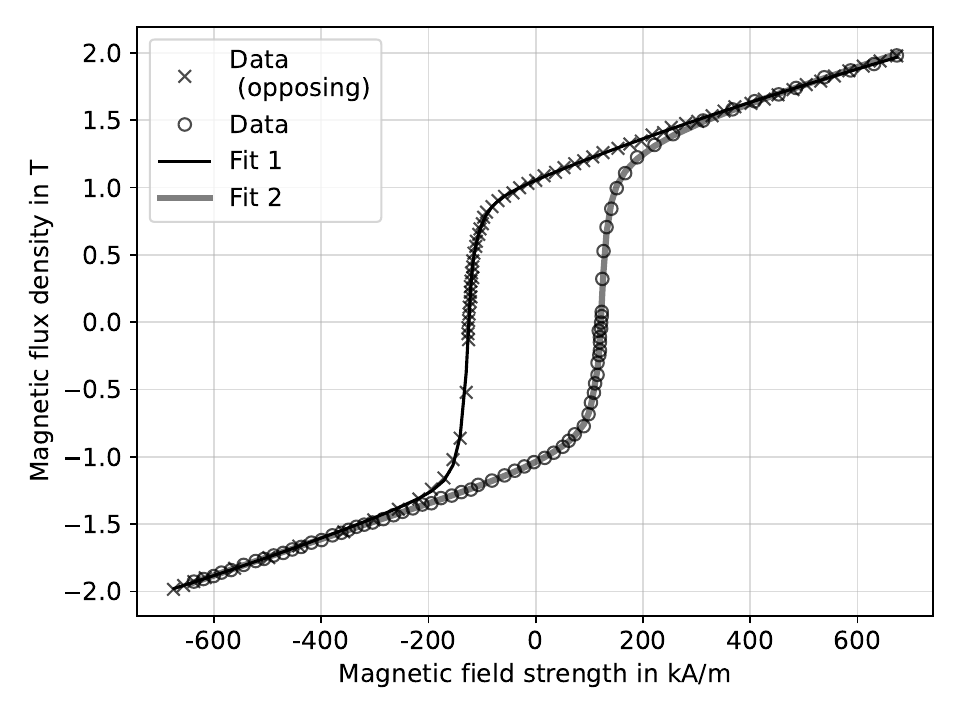}
		\caption{Hysteresis behavior of Alnico 9 LNGT72 \cite{alnico} fitted with 5 S-curves.}
		\label{fig7b}
	\end{subfigure}
	\caption{Hysteresis behavior of Alnico \cite{hyster_data}. (a) The estimated end points of hysteresis loop are ($-348.64$ kA/m,$-1.551$ T) and (334.99 kA/m, 1.54 T) that enclose an area of 515.69 kJ.(b) The estimated end points of hysteresis loop are ($-605.215$ kA/m,$-1.89$ T) and (481.39 kA/m, 1.735 T) that enclose an area of 550.8 kJ.}
	\label{fig7}
\end{figure}
Magnetization not only depends on th applied magnetic field but also on the previous states of magnetization \cite{mdpi_23}. Thus, with the alternating applied field, ferromagnetic substances exhibit hysteresis.
We now represent hysteresis loop using Eqn. (\ref{eq_contfrac}) as shown in Fig. (\ref{fig4a}). The upper and lower S-curves of the hysteresis loop are obtained by shifting the point of inflection while keeping parameters $a$ and $m$ constant. Fig. (\ref{fig4a}) demonstrates that it is possible to obtain a representation of hysteresis with just two parameters and two points of inflection.
The shaded area in Fig. (\ref{fig4a}) is given by
\begin{equation}
\nonumber \int_{x_1}^{x_2}[y(x)_{(-4.4,13)}-y(x)_{(6.4,26)}]dx=\int_{y_1}^{y_2}x(y)_{(-4.4,13)}-x(y)_{(6.4,26)}dy
\end{equation}
The area can be calculated in a straight-forward manner along the $y-$ axis. The area in Fig. (\ref{fig4a}) is found to be 1033.57 J.

Similar to Fig. (\ref{fig4a}), the hysteresis data of a soft ferrite substance Mn$_{0.51}$Zn$_{0.44}$Fe$_{2.05}$O$_4$ \cite{maganalyst,anhys5} is considered in Fig. (\ref{fig4b}). However, since we are considering real data, the dissipation component is included and the superposition model fits well with the data. The upper S-curve (shown as `Fit 1') in Fig. (\ref{fig4b}) is approximated as
\begin{equation}
y_{\text{net}}^{(\text{upper})}=-1.673y(x,2.873,0.004,-8.4682,-0.0035)+1.55y(x,2.873,0.0106,-11.796,0),
\label{eq_upper}
\end{equation}
where $p_1m_1=-0.0067$ and $p_2m_2=0.0165$. The lower curve (shown as `Fit 2') is approximated as
\begin{equation}
y_{\text{net}}^{(\text{lower})}=-1.655y(x,2.804,0.004,8.172,-0.0365)+1.53y(x,2.804,0.011,11.5,0),
\label{eq_lower}
\end{equation}
where $p_1m_1=-0.0069$ and $p_2m_2=0.0167$.
In both the above approximations, the first terms represent dissipation and the second terms represent magnetizations in the respective directions. The maximum permeability for $y_{\text{net}}^{(\text{upper})}$ is 0.00976 H/m at applied magnetic intensity $-11.99$ A/m and for $y_{\text{net}}^{(\text{lower})}$ it is 0.0099 H/m at 11.698 A/m.

The intersection of the two curves $y_{\text{net}}^{(\text{upper})}$ and 
$y_{\text{net}}^{(\text{lower})}$ can be obtained using root-finding techniques (Sec (\ref{sec_newton})) by solving Eqns. (\ref{eq_upper}) and (\ref{eq_lower}). This helps us to find the area of the hysteresis loop. In Fig. (\ref{fig4b}), the area of loop is found to be 14.74 J. As shown in the previous section, the upper and lower S-curves can further be expressed as sums of corresponding subprocesses.

Similarly, we fit various publicly available hysteresis data \cite{maganalyst,hyster_data}. For example, in Fig. (\ref{fig5}), hysteresis loops of soft magnetic substances are considered and good fits are obtained using S-curve superposition. In Figs. (\ref{fig6}) and (\ref{fig7}), good fits are obtained for hysteresis data of Alnico alloys.
\section{Profiling data}
The nonlinear regression model using continued fraction of straight lines can be used to fit, interpret and profile data. So far, the literature has only the maximum permeability as a reliable measure to describe a basic magnetization curve \cite{magweb_book}. In this work, we introduce another measure to describe its nonlinearity.

Using Eqn. (\ref{eq_contfrac}) in its algebraic form we get
\begin{eqnarray}
\frac{dy}{dx}&=&\frac{m}{1+3a(y-y_c)^2}\\
\frac{d^2y}{dx^2}&=&-\frac{6a(y-y_c)}{1+3a(y-y_c)^2}\left(\frac{dy}{dx}\right)^2\\
\frac{d^3y}{dx^3}&=&-\frac{6a}{1+3a(y-y_c)^2}\left(\frac{dy}{dx}\right)
{\left(\frac{dy}{dx}\right)^2+3(y-y_c)\left(\frac{d^2y}{dx^2}\right)}
\end{eqnarray}
In the above expression, $dy/dx$ gives us the maximum permeability in a magnetization curve $a(y-y_c)^3+(y-y_c)=m(x-x_c)$. ${d^2y}/{dx^2}$ helps us to locate the inflection point on the curve. Similarly, definite findings are possible on magnetization data approximated by the S-curve superposition $y_{\text{net}}$. So far, for profiling data with this method, the following measures have been proposed \cite{twosupS}:
\begin{enumerate}
\item Percentage nonlinearity which is given by
\begin{equation}
\frac{|\sum_{n=0}^{n-1}p_im_i-m|}{m},
\end{equation}
where $m$ is obtained from the following equation
\begin{equation}
\frac{d^2y_{\text{net}}}{dx^2}=0.
\end{equation}
$m$ is therefore, the maximum slope of the monotonically rising $y_{\text{net}}$ curve. $m$ is also the maximum permeability.
\item The measure $m/(1+a)$, where $a$ is a parameter of $y_{\text{net}}$.
As $a\rightarrow 0$, the above measure becomes $m$ and for large $a$ the measure diminishes. This incorporates the influence of $a$.
\end{enumerate}
However, all the parameters other than $m$ in the above measures are sensitive to initial conditions \cite{twosupS}. The superposition model provides definite measure of $m$ and its location i.e., the point of inflection. On the other hand, the two-parameter S-curve $a(y-y_c)^3+y-y_c=m(x-x_c)$ (fitted in Fig.(\ref{fig1a_am})) is robust to initial conditions (see Table 9 of \cite{twosupS} and Table II of \cite{shruti_Ssum}). 

We will use the two-parameter S-curve to quantify the region around the inflection point where magnetization of initially aligned domains reaches a maximum and dissipation due to domain wall motion reaches a minimum.
This is an ideal state of the magnetization process because:
\begin{enumerate}
\item Initially aligned domains have reached their maximum magnetization and they have not yet started dissipating.
\item Rest of the domains align rapidly with minimum dissipation due to domain wall motion. 
\end{enumerate}
Since the point of inflection represents an ideal state with minimal disspation, we refer to the point as the origin, say $(x_0,y_0)$ with parameters $a_0$ and $m_0$. At the origin, magnetization occurs with maximum permeability in the ferromagnetic substance. The two-parameter S-curve is now represented as
\begin{equation}
a_0(y-y_0)^3+y-y_0=m_0(x-x_0).
\label{eq_profile_2S}
\end{equation}
\subsection{Estimation using Newton-Raphson method}
\label{sec_newton}
The Newton-Raphson method has been used to find roots almost up to machine precision \cite{jog2}. Fitting $y_{\text{net}}$ on magnetization data, we can precisely identify $(x_0,y_0)$ using the Newton-Raphson method by solving
\begin{eqnarray}
\nonumber\frac{d^2y_{\text{net}}}{dx^2}&=&0\\
\Rightarrow \sum_{i=0}^{n-1}p_i\frac{d^2y_i}{dx^2}&=&0,
\end{eqnarray}
where $y_i=y(a,m_i,x-x_{ci},y_{ci})$. The slope of $y_{\text{net}}$ at $(x_0,y_0)$ gives us the precise estimation of $m_0$.

\begin{figure}[ht!]
	\centering
	\begin{subfigure}[b]{0.45\textwidth}
		\includegraphics[width=\textwidth]{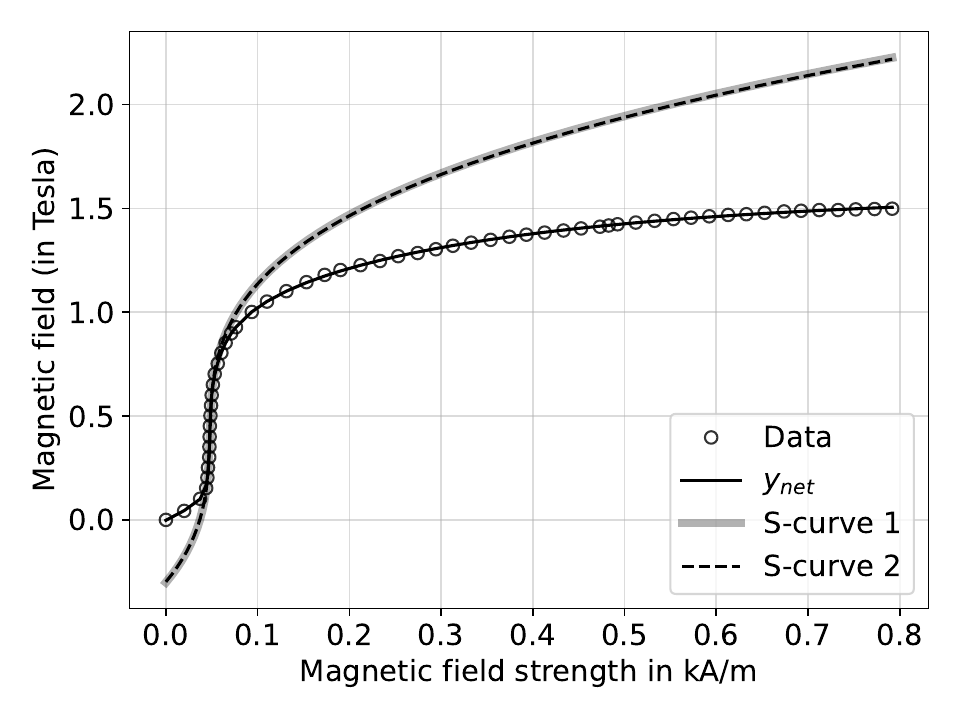}
		\caption{$m_0=0.16301014590032463 \text{ H/m}$ and $a_0\in[19.88328004306248, 20.113215614093658] \text{ T}^{-2}$. }
		\label{fig8a_am}
	\end{subfigure}
	\begin{subfigure}[b]{0.45\textwidth}
		\includegraphics[width=\textwidth]{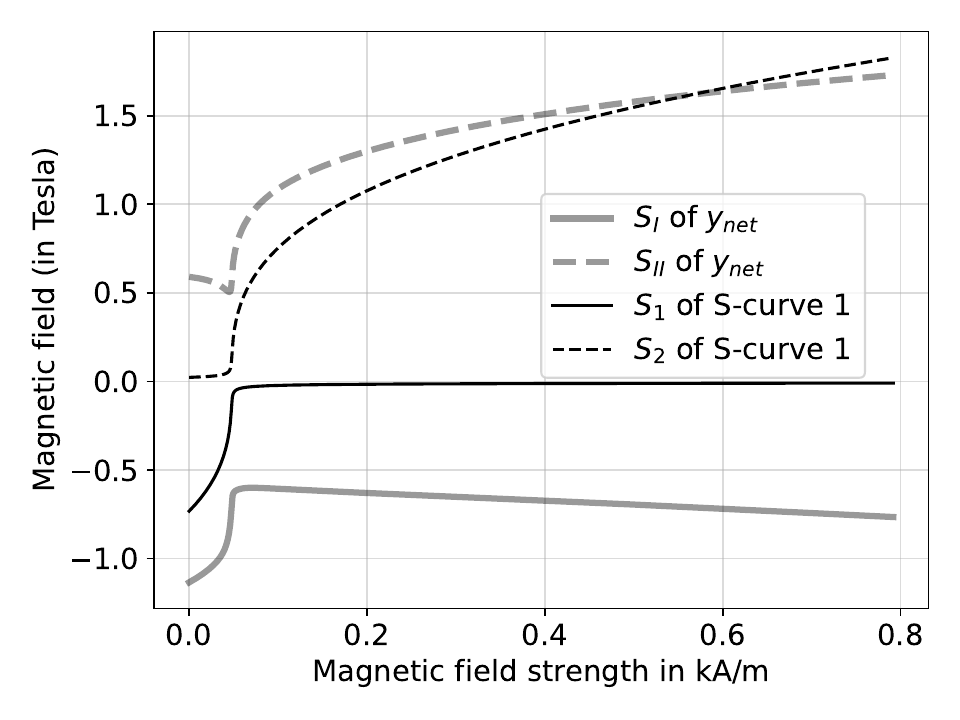}
		\caption{The two subprocesses of `S-curve 1' with minimal dissipation representing an ideal process.}
		\label{fig8b_amsum}
	\end{subfigure}
	\caption{S-curve superposition provides accurate estimation of inflection point $(47.771706421976784 \text{ H/m}, 0.4078417069999283 \text{ T})$ and permeability $m$ at that point using the Newton-Raphson method. Parameter $a$ varies around an interval reflecting asymmetrical nature of $y_\text{net}$ unlike the ideal S-curve which is symmetric around the inflection point. `S-curve 1' and `S-curve 2' almost lie on top of each other.}
	\label{fig8_basic0}
\end{figure}

We will now estimate $a_0$ assuming symmetric deviation from linearity around the origin $(x_0,y_0)$. Around any inflection point, the second order derivative $d^2y/dx^2$ reaches extreme values, say, at $x_1$ and $x_2$. Therefore, we use the condition
\begin{eqnarray}
\frac{d^3y_\text{net}}{dx^3}=0.
\end{eqnarray}
Precise values of $x_1$ and $x_2$ are obtained by solving the above equation numerically again using the Newton-Raphson method. We will then have the following equations for an ideal representation of the two parameter S-curves
\begin{eqnarray}
\nonumber a_0(y_1-y_0)^3+y_1-y_0&=&m_0(x_1-x_0),\\
\nonumber a_0(y_2-y_0)^3+y_2-y_0&=&m_0(x_2-x_0),
\end{eqnarray}
where $y_1=y_{\text{net}}(x_1)$ and $y_2=y_{\text{net}}(x_2)$. If $y_{\text{net}}$ is symmetric around $(x_0,y_0)$, then $a_0$ is a real number as in Figs. (\ref{fig1a_am}) and (\ref{fig4a}) or Fig. 9 of \cite{a_m_mod}. Otherwise, for $x\in[x_1.x_2]$, $a_0$ will be obtained as a range of values, $a_0\in[a_1,a_2]$. Narrower the range more ideal a magnetization process is. In Fig. (\ref{fig8a_am}), `S-curve 1' represents S-curve with $a_0=a_1$ and `S-curve 2' represents S-curve with $a_0=a_2$.

The ideal $(a_0,m_0)$ curve is symmetric and the subprocess saturates with minimal dissipation as shown in Fig. (\ref{fig8b_amsum}) (similar to Fig. 9 of \cite{a_m_mod}). The Newton-Raphson method is also useful in finding the intersection of upper and lower S-curves of a hysteresis loop (Figs. (\ref{fig4})-(\ref{fig7})).
\subsection{Summary}
The procedure of fitting and estimation of parameters is summarized below:
\begin{enumerate}
\item Eqn. (\ref{eq_ynet}) is used to fit the magnetization data as a nonlinear regression model. Parameters such as $p_i'$s, $m_i'$s and $a$ are estimated based on the least squares fitting of the model. $x_{ci}'$s and $y_{ci}'$s are chosen from the data points following the procedure in \cite{twosupS,shruti_Ssum}.

\item Based on the fitted parameters, $p_im_i$ can be positive or negative. Positive components represent magnetization while the negative components represent dissipation. 

\item The fitted model can be expressed as a sum of two subprocesses. The subprocesses represent two kinds of mechanisms in the magnetization process. They are, magnetization of domains that are already aligned with the applied field, and magnetization of domains with misaligned magnetic moments. 

\item Although the model fits well for the magnetization and hysteresis data of both soft and hard ferromagnetic substances, the parameters $p_i'$s, $m_i'$s and $a$ are sensitive to initial values (i.e. the starting guess). Hence, in order to obtain definite measures, we use representative two-parameter S-curves (Eqn. \ref{eq_am_s}) expressed as $a_0(y-y_0)^3+y-y_0=m_0(x-x_0)$. The $(a_0,m_0)$ curve represents the region of maximum magnetization and the values can be obtained almost up to machine precision using the Newton-Raphson method.
\end{enumerate}
\subsection{Profiling basic magnetization curves}
In this section, we profile various publicly available basic magnetization curves provided by MagWeb \cite{magweb0} by identifying their inflection points $(x_0,y_0)$, maximum permeability $m_0$ and the interval of $a_0$. The results are shown for some of the materials in Table \ref{tab1} and in Fig. (\ref{fig9}) provided in the appendix.

The last three grades in Table \ref{tab1} contain highest permeabilities. PC40 reaches inflection point with just 10.82 A/m of applied field strength, whereas 50PN600 requires 99.11 A/m to reach $y_0$. Moreover, induced flux density of PC40 is 0.089 T which is far less than the other two grades. The $a-$ interval indicates that the magnetization of 50PN600 and 50CS600 is more symmetrical about $(x_0,y_0)$ indicating minimal dissipation. The values of $a_0-$ interval indicates nonlinearity. For example, the $a_0-$ values of 50CS600 are less than those of 50PN600. Therefore, 50PN600 reaches saturation with less applied field strength than required for 50CS600. 

Next, we consider magnetization data of grade `M-19 14mil' provided for various frequencies of the applied field. The representative parameters of the magnetization processes are shown in Table \ref{tab2} and in Fig. (\ref{fig10}). $m_0$ decreases with increase in frequency and so does $y_0$ even for large $x_0$. The $a_0-$ interval is almost a single value at 200Hz. The 200Hz plot in Fig. (\ref{fig10}) is the most symmetric curve among all the other curves. 
We are unable to estimate the $a_0-$ interval at 1000Hz because both extremes of $d^y_{\text{net}}/dx^2$ does not lie within the data. The $a_0-$ interval for 2000Hz is not estimated accurately because of the limited number of points. This is also the reason we do not get a presentable fit for grade `2605SA1 .025mm' data. This is a limitation of the superposition model $y_{\text{net}}$.

Table \ref{tab3} and Fig. (\ref{fig11}) show estimated parameters for magnetization data of grade PC40 at various temperatures. Although it can be observed that $m_0$ decreases with the increase in temperature. $m_0$ at 120 C shows a slight increase in permeability. However, the induced field $y_0$ is higher at 100 C than at 120 C for nearly the same $x_0$. Magnetization curves at 60 C and 120 C are highly asymmetrical about $(x_0,y_0)$ as shown by the $a_0-$ interval.
\subsection{Profiling demagnetization curves}
When a magnet is operated upon beyond its `knee point' in a demagnetizing curve, the magnet demagnetizes \cite{demag1}. While there are many definitions of `knee point', in this work, we look for the point of maximum curvature or maximum value of $|d^2y_{\text{net}}/dx^2|$ in a demagnetizing curve. This point is given by $(x_k,y_k)$ in Table \ref{tab4} and also in Fig. (\ref{fig12}). We cannot find $a_0-$ interval as the demagnetizing curves do not contain extreme values of $|d^2y_{\text{net}}/dx^2|$ at both sides of the inflection point.
In Table \ref{tab4}, NMX-48BH at 100C and 140C have high $m_0$ and $y_k$ values. NMX-48BH at 140 C demagnetizes with less applied field strength $x_k$ than at 120 C.
\section{Conclusions}
A variety of nonlinear magnetization data including hysteresis loop has been fitted using the superposition of continued fraction of straight lines i.e. two-parameter S-curves. The superposition is a combination of permeable (positive slopes) and dissipative components (negative slopes), and it is a sum of two distinct subprocesses of magnetization. One of the subprocesses is magnetization of domains that are aligned with the externally applied field. The other subprocess is the magnetization of domains that align only at high strength of the applied field. Around the inflection point, all the domains align rapidly with the increase in applied field and the process exhibits maximum net permeability. This permeability reduces with further increase in applied field strength until saturation. Further, we have profiled the magnetization data using the parameters obtained from the fitted model. Based on the profiles, the symmetry and nonlinearity of magnetization curves can be described. Symmetry indicates less dissipation and nonlinearity indicates early saturation. Thus, this paper demonstrates that the magnetization of ferromagnetic substances can be described using smooth algebraic equations that represent S-curves, instead of special constructs such as exponential or trigonometric functions. 
\section{Code and supplementary material}
	The code and the supplementary material are provided in the github respository: grasshopper14/Continued-fraction-of-straight-lines within magnetization folder.
	\bibliographystyle{unsrt}

\section*{Appendix}
	\begin{table}[h!]
		\centering
		\begin{tabular}{|c|c|c|c|c|c|}
			\hline
			code& Grade & $a_0-$ interval in T$^{-2}$& $m_0$ in H/m & $x_0$ in A/m & $y_0$ in T \\\hline
			1& B50A600 & [1.346647,1.658100] & 0.012482 & 85.832314 & 0.491577 \\\hline
			2& Koolmu 40mu & [2.225875, 2.313996] & 0.000056 & 937.457608 & 0.061355 \\\hline
			3& 1020 Steel Annealed & [0.374910,1.991545] & 0.002033 & 84.498903 & 0.175003 \\\hline
			4& 500 1P 600MPa & [1.493642,1.648367] & 0.000733 & 476.287937 & 0.283668 \\\hline
			5&Cast Iron& [3.386896, 3.413355] & 0.002355 & 146.328607 & 0.274111 \\\hline
			6& 430-FR 15mm & [0.624275, 1.943880] & 0.003412 & 228.621129 & 0.561336 \\\hline
			7& PC40 & [69.756087, 85.400248] & 0.013539 & 10.819921 & 0.088543 \\\hline
			8& 50PN 600 & [1.214966,1.475583] & 0.011858 & 99.111978 & 0.542716 \\\hline
			9& 50CS600 & [0.766845,0.854387] & 0.013574 & 76.886035 & 0.489805 \\\hline
		\end{tabular}
		\caption{Representative measures to profile magnetization data based on the fits provided in Fig. (\ref{fig9}). The numbers in the `Code' column of the above table correspond to the labelled subplots in Fig. (\ref{fig9}).}
		\label{tab1}
	\end{table}
	\begin{figure}[ht!]
		\centering
		\includegraphics[width=0.9\textwidth]{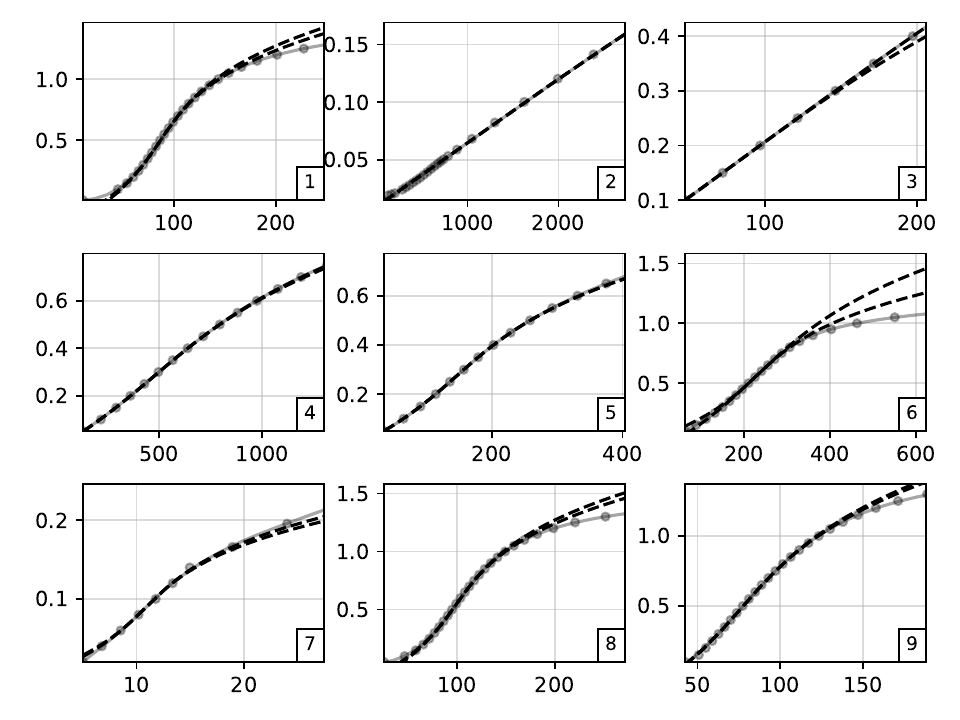}
		\caption{Fitting S-curve superposition on magnetization data \cite{magweb0} provided for various grades of magnetic materials. $m_0/(4\pi\times 10^{-7})$ provides the maximum relative permeability. They are: 1 - 9932.85, 2 - 44.56, 3 - 1617.81, 4 - 583.30, 5 - 1874.05, 6 - 2715.18, 7 - 10773.99, 8 - 9436.2, 9 - 10801.84. Faded dots and lines are data points and fits, respectively. The dashed dark lines are representative S-curves for different $a_0$ values: $a_1$ and $a_2$.}
		\label{fig9}
	\end{figure}
\begin{table}[h!]
	\centering
	\begin{tabular}{|c|c|c|c|c|}
		\hline
		Frequencies & $a_0-$interval in T$^{-2}$ & $m_0$ in H/m & $x_0$ in A/m & $y_0$ in T \\\hline
		0 Hz& [0.914666, 1.700438] & 0.013149 & 52.360314 & 0.513765 \\\hline
		50 Hz& [1.357840, 2.255541] & 0.013810 & 58.309731 & 0.490997 \\\hline
		60 Hz & [1.470959, 2.634390] & 0.014769 & 55.965253 & 0.442782 \\\hline
		100 Hz & [1.121324, 2.229939] & 0.012784 & 60.594651 & 0.470565 \\\hline
		150 Hz & [0.785626, 0.904428] & 0.011264 & 73.224892 & 0.556770 \\\hline
		200 Hz & [0.399404, 0.399404] & 0.009229 & 59.939935 & 0.387719 \\\hline
		300 Hz & [0.744969, 0.819192] & 0.008982 & 92.287758 & 0.577706 \\\hline
		400 Hz & [0.803348, 0.830091] & 0.009079 & 92.800607 & 0.582042 \\\hline
		600 Hz & [2.746700,2.988539] & 0.009655 & 137.753784 & 0.781576 \\\hline
		1000 Hz & - & 0.005391 & 119.205483 & 0.487006 \\\hline
		2000 Hz & [18.461876, 18.516705] & 0.004274 & 174.634418 & 0.476462 \\\hline
		4000 Hz & [8.058052, 11.513848] & 0.002495 & 143.192575 & 0.267568 \\\hline
	\end{tabular}
	\caption{Representative measures to profile magnetization data at various frequencies of the applied field based on the fits provided in Fig. (\ref{fig10}). These values correspond to the grade `M-19 14mil' \cite{magweb0}.}
	\label{tab2}
\end{table}
\begin{figure}[ht!]
	\centering
	\includegraphics[width=0.9\textwidth]{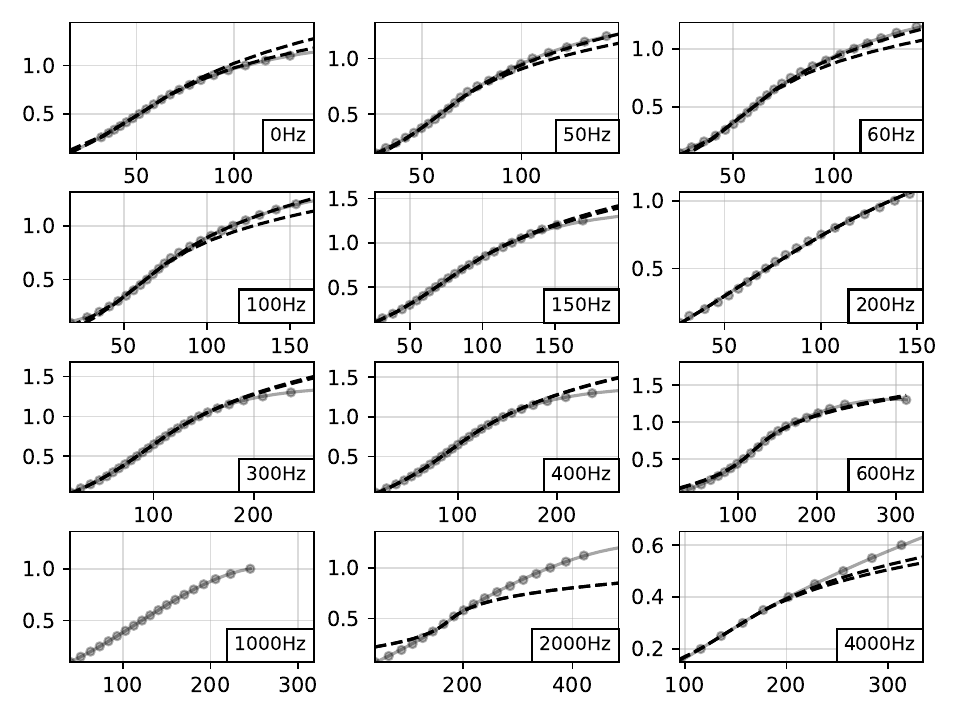}
	\caption{Fitting S-curve superposition on magnetization data of grade `M-19 14mil' \cite{magweb0} at various frequencies of the applied magnetic field. $m_0/(4\pi\times 10^{-7})$ provides the maximum relative permeability. They are: 0Hz - 10463.64, 50Hz - 10989.6, 60Hz - 11752.8, 100Hz - 10173.18, 150Hz - 8963.6, 200Hz - 7344.2, 300Hz - 7147.64, 400Hz - 7224.83, 600Hz - 7683.2, 1000Hz - 4290.02, 2000Hz - 3401.14, 4000Hz - 1985.45.  $x-$ axis represents applied magnetic field stright in A/m and $y-$ axis represents induced magnetic field in T. Faded dots and lines are data points and fits, respectively. The dashed dark lines are representative S-curves for different $a_0$ as $a_1$ and $a_2$ values. }
	\label{fig10}
\end{figure}
\begin{table}[h!]
	\centering
	\begin{tabular}{|c|c|c|c|c|}
		\hline
		Temperature & $a_0-$interval in T$^{-2}$ & $m_0$ in H/m & $x_0$ in A/m & $y_0$ in T \\\hline
		25 C & [69.756087, 85.400248] & 0.013539 & 10.819921 & 0.088543 \\\hline
		60 C & [15.373073, 195.255141] & 0.006272 & 6.16 & 0.020873 \\\hline
		100 C & [237.514995, 270.686598] & 0.006061 & 36.028006 & 0.147021 \\\hline 
		120 C & [99.424598, 165.749843] & 0.007201 & 36.669506 & 0.105648 \\\hline
	\end{tabular}
	\caption{Representative measures to profile magnetization data at various temperatures based on the fits provided in Fig. (\ref{fig11}). These values correspond to the grade `PC40' \cite{magweb0}.}
	\label{tab3}
\end{table}
\begin{figure}[ht!]
	\centering
	\includegraphics[width=0.8\textwidth]{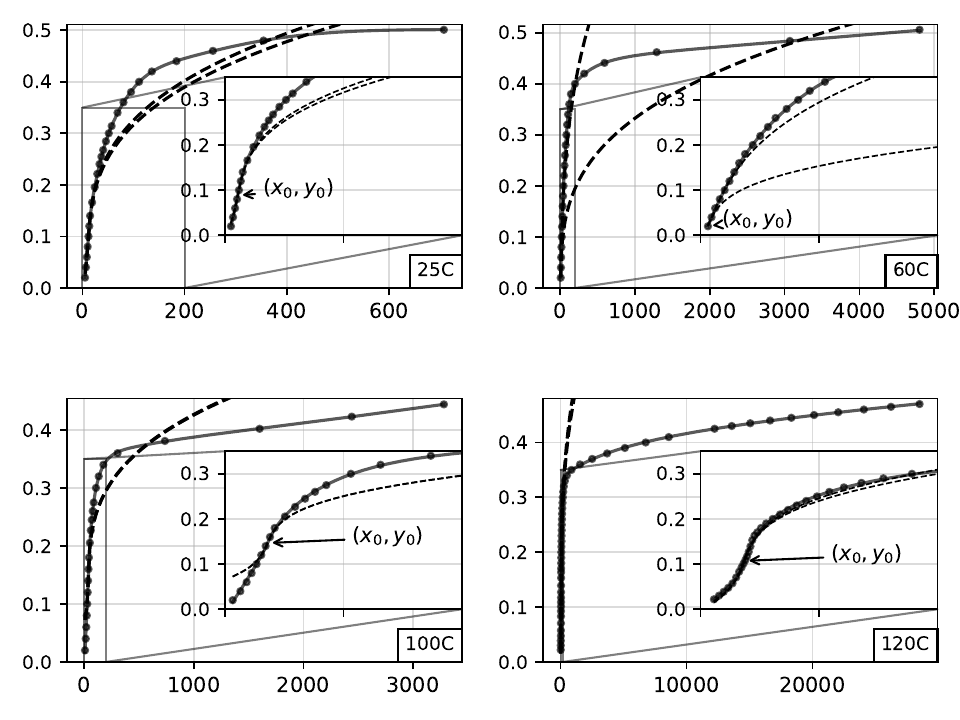}
	\caption{Fitting S-curve superposition on magnetization data of grade `PC40' \cite{magweb0} at various temperatures. Insets show the range of applied magnetic field strength for which change in magnetization is maximum around point of inflection $m_0/(4\pi\times 10^{-7})$ provides the maximum relative permeability. They are: 25 C - 10773.99, 60C - 4991.1, 100C - 4823.19, 120C - 5730.37. $x-$ axis represents applied magnetic field stright in A/m and $y-$ axis represents induced magnetic field in T.}
	\label{fig11}
\end{figure}
\begin{table}[h!]
	\centering
	\begin{tabular}{|c|c|c|c|c|c|c|}
		\hline
		Code&Grade & $m_0$ in H/m & $x_0$ in A/m & $y_0$ in T& $x_k$ in A/m& $y_k$ in T \\\hline
		1&LNGT18 at 80C & 0.009164 & -102.094368 & 0.044985 & -95.639154 & 0.102602 \\\hline
		2&S3218 at 20C & 0.001427 & -750.722620 & 0.154365 & -711.303630 & 0.210360 \\\hline
		3&S3218 at 200C & 0.004418 & -1474.500747 & -2.708491 & -745.700000 & -0.000256 \\\hline
		4&N40UH at 150C & 0.003215 & -735.172387 & 0.020241 & -705.204710 & 0.110482 \\\hline
		5&NMX-48BH at 60C & 0.066302 & -711.000000 & -0.006954 & -707.231323 & 0.211753 \\\hline
		6&NMX-48BH at 100C& 0.118917 & -443.415362 & 0.280585 & -441.093509 & 0.515612 \\\hline
		7&S3218 at 300C & 0.014517 & -565.671937 & -0.430003 & -517.958696 & 0.161832 \\\hline
		8&NMX-48BH at 140C& 0.332164 & -323.470700 & 0.432717 & -322.950405 & 0.583386 \\\hline
		9&S3218 at 100C & 0.001624 & -800.522052 & 0.036497 & -790.556706 & 0.052564 \\\hline
		10&LNGT18 at 150C & 0.011789 & -88.415342 & 0.046876 & -87.824342 & 0.053663 \\\hline
		11&Alnico 9 at 20C& 0.064139 & -121.362317 & 0.011226 & -117.648065 & 0.234950 \\\hline
		12&HF085 at 20C& 0.013526 & -1023.076322 & -7.645509 & -262.598903 & 0.044988 \\\hline
		13&2560 at $-40$C& 0.010405 & -194.150324 & 0.023387 & -193.637924 & 0.028413 \\\hline
		14&86EP at 100C& 0.004062 & -340.711789 & 0.038924 & -328.849685 & 0.083173 \\\hline
		15&N40UH at 180C& 0.037707 & -530.312079 & 0.051122 & -528.701270 & 0.104681 \\\hline
		16&86EP at 80C & 0.002952 & -411.506810 & -0.081228 & -360.832883 & 0.058367 \\\hline
		17&FB138 at $-60$C & 0.013405 & -319.560528 & 0.010875 & -314.222622 & 0.073628 \\\hline
	\end{tabular}
	\caption{Representative measures of fits shown in Fig. (\ref{fig12}) to profile demagnetization data \cite{magweb0}.$(x_k,y_k)$ are the points of maximum curvature.}
	\label{tab4}
\end{table} 
\begin{figure}[ht!]
	\centering
	\includegraphics[width=1\textwidth]{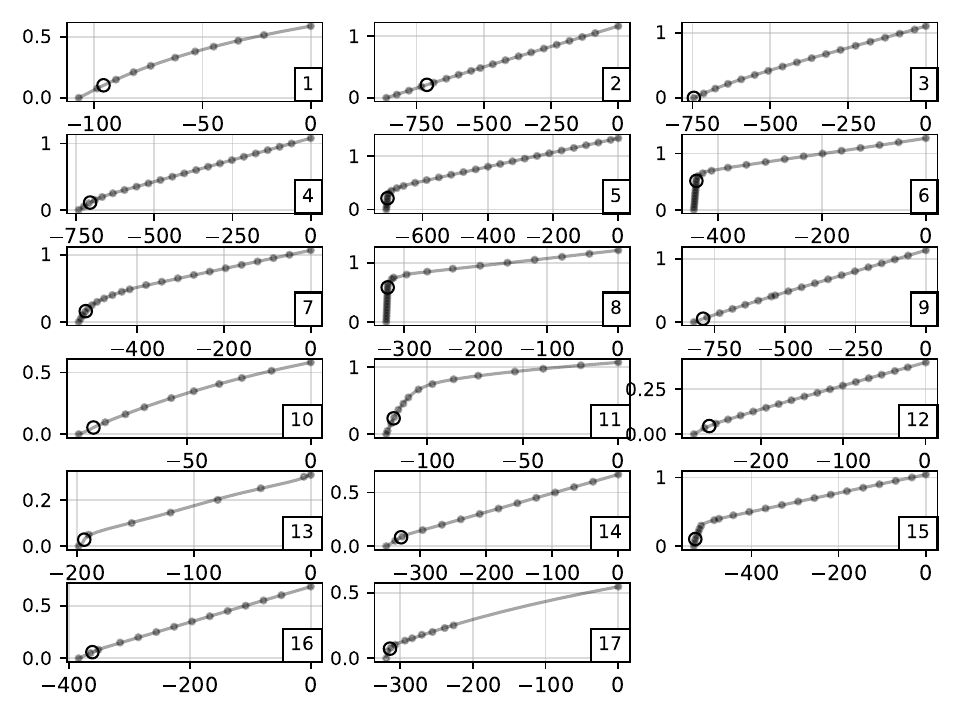}
	\caption{Fitting S-curve superposition on demagnetization data.  $x-$ axis represents applied magnetic field stright in A/m and $y-$ axis represents induced magnetic field in T. Faded dots and lines are data points and fits, respectively. `$\circ$' is the point with maximum curvature $(x_k,y_k)$.}
	\label{fig12}
\end{figure}
\end{document}